%% file: main.tex
\documentclass[10pt,twocolumn,letterpaper]{article}

\usepackage[pagenumbers]{cvpr}
\usepackage{epsfig}

\usepackage{graphicx}
\usepackage{amsmath}
\usepackage{amssymb}
\usepackage{booktabs}
\usepackage[accsupp]{axessibility}
\usepackage{multirow}
\usepackage{color}
\usepackage{pifont}
\usepackage[misc,geometry]{ifsym} 
\renewcommand{\paragraph}[1]{\vspace{1mm}\noindent\textbf{#1}\hspace{1em}}

\definecolor{linkcolor}{RGB}{255,0,0}
\definecolor{urlcolor}{RGB}{255,105,180}
\definecolor{citecolor}{RGB}{66,168,235}
\usepackage[pagebackref,breaklinks,colorlinks]{hyperref}
\hypersetup{colorlinks=true,linkcolor=linkcolor,urlcolor=urlcolor,citecolor=citecolor}

\usepackage[capitalize]{cleveref}
\crefname{section}{Sec.}{Secs.}
\Crefname{section}{Section}{Sections}
\Crefname{table}{Table}{Tables}
\crefname{table}{Tab.}{Tabs.}

\makeatletter
\DeclareRobustCommand\onedot{\futurelet\@let@token\@onedot}
\def\@onedot{\ifx\@let@token.\else.\null\fi\xspace}
\def\@fnsymbol#1{\ensuremath{\ifcase#1\or \textsuperscript{~\Letter}\or \ddagger\or
   \mathsection\or \mathparagraph\or \|\or **\or \dagger\dagger
   \or \ddagger\ddagger \else\@ctrerr\fi}}
\makeatother

\def\etal{\emph{et al}\onedot}
\makeatother

\def\cvprPaperID{426}
\def\confName{CVPR}
\def\confYear{2023}

\begin{document}

\title{Panoptic Video Scene Graph Generation}
\author{
Jingkang Yang$^{*,\dagger}$, Wenxuan Peng$^{*,\dagger}$, Xiangtai Li$^\dagger$, Zujin Guo$^\dagger$, Liangyu Chen$^\dagger$, Bo Li$^\dagger$\\ 
Zheng Ma$^\ddagger$, Kaiyang Zhou$^\dagger$, Wayne Zhang$^\ddagger$, Chen Change Loy$^\dagger$, Ziwei Liu$^\dagger$\footnote{Corresponding author}
\and
$^\dagger$S-Lab, Nanyang Technological University, Singapore\\
$^\ddagger$SenseTime Research, Shenzhen, China\\
{\tt\small\url{https://jingkang50.github.io/PVSG}}
}

\twocolumn[{%
   \renewcommand\twocolumn[1][]{#1}%
   \maketitle
   \vspace{-38pt}
   \begin{center}
    \centering
    \includegraphics[width=0.94\linewidth]{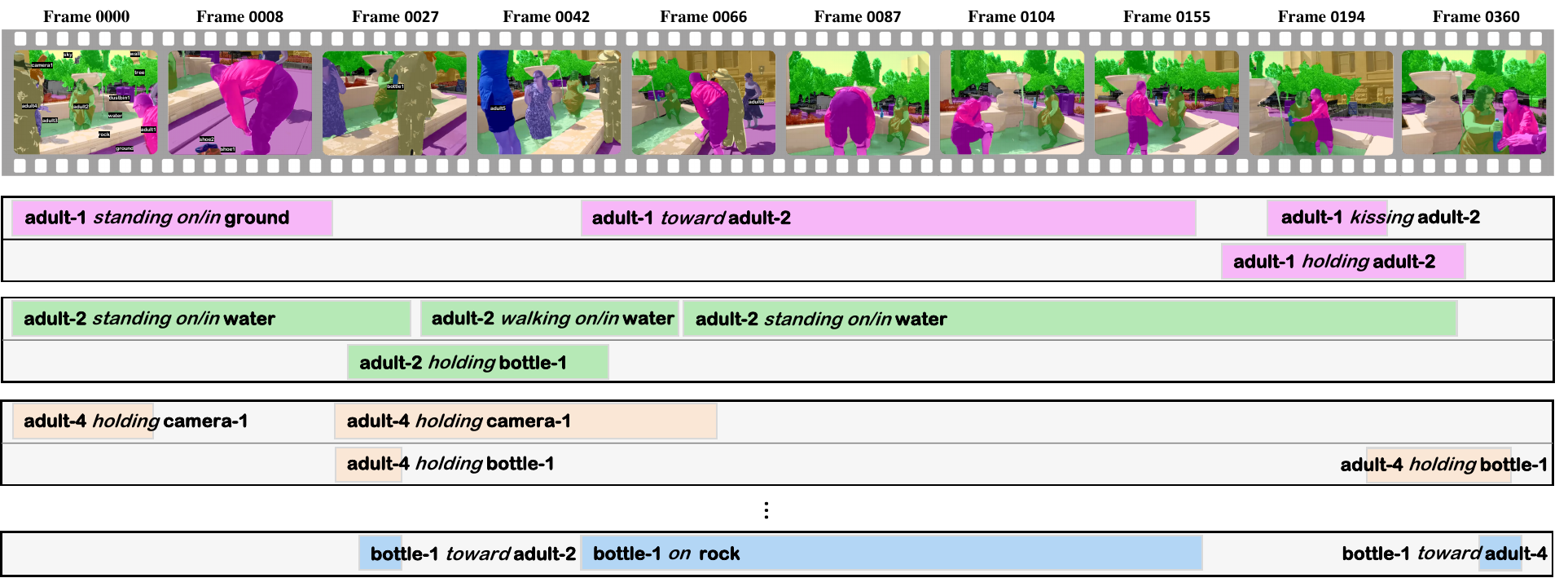}
    \vspace{-5pt}
    \captionof{figure}{
    \textbf{An example video from our panoptic video scene graph (PVSG) dataset}. The top row shows some keyframes overlaid with the frame-wise panoptic segmentation masks. The timeline tubes underneath the keyframes contain fine, temporal scene graph annotations. The PVSG dataset contains 400 videos (with an average duration of 76.5 seconds), including 289 third-person and 111 egocentric videos.
    }
    \label{fig:teaser}
   \end{center}%
  }]
  \def\thefootnote{*}\footnotetext{Main contributors.
  \textsuperscript{~\Letter} Corresponding author.}
\def\thefootnote{\arabic{footnote}}

\input{section/0_abstract.tex}
\input{section/1_introduction.tex}
\input{section/2_related_work.tex}
\input{section/3_problem.tex}
\input{section/4_dataset.tex}

\input{section/5_baselines.tex}

\input{section/6_results.tex}
\input{section/7_discussion.tex}

{\small
\bibliographystyle{ieee_fullname}
\bibliography{egbib}
}

\newpage
\input{section/8_appendix.tex}

\end{document}

%% file: section/0_abstract.tex
\begin{abstract}
Towards building comprehensive real-world visual perception systems, we propose and study a new problem called panoptic scene graph generation (PVSG). PVSG relates to the existing video scene graph generation (VidSGG) problem, which focuses on temporal interactions between humans and objects grounded with bounding boxes in videos. However, the limitation of bounding boxes in detecting non-rigid objects and backgrounds often causes VidSGG to miss key details crucial for comprehensive video understanding. In contrast, PVSG requires nodes in scene graphs to be grounded by more precise, pixel-level segmentation masks, which facilitate holistic scene understanding.
To advance research in this new area, we contribute the PVSG dataset, which consists of 400 videos (289 third-person + 111 egocentric videos) with a total of 150K frames labeled with panoptic segmentation masks as well as fine, temporal scene graphs. We also provide a variety of baseline methods and share useful design practices for future work.
\end{abstract}

%% file: section/1_introduction.tex
\section{Introduction}
\label{sec:intro}

\begin{table*}[t]
\caption{\textbf{Comparison between the PVSG dataset and some related datasets}. Specifically, we choose three video scene graph generation (VidSGG) datasets, three video panoptic segmentation (VPS) datasets, and two egocentric video datasets---one for short-term action anticipation (STA) while the other is for video object segmentation (VOS), for comparison. Our PVSG dataset is the first long-video dataset with rich and fine-grained annotations of panoptic segmentation masks and temporal scene graphs.}
\label{tab:dataset}
\vspace{-8pt}
\centering
\setlength\tabcolsep{5pt}
\resizebox{\textwidth}{!}{
\begin{tabular}{@{\hskip 0.05in}l@{\hskip 0.2in}cccccccccccccc@{\hskip 0.05in}}
\toprule
Dataset &Task & \#Video & Video Hours & Avg. Len. & View & \#ObjCls & \#RelCls & Annotation & \# Seg Frame & Year & Source\\ \midrule
ImageNet-VidVRD~\cite{shang2017video}  & VidSGG & 1,000  &  -  & - &  3rd   & 35  & 132 & Bounding Box  & - & 2017 &  ILVSRC2016-VID~\cite{russakovsky2015imagenet} \\ 
Action Genome~\cite{ji2020actiongenome}    & VidSGG & 10,000  &  99  & 35s &  3rd   & 80  & 50 & Bounding Box  & - & 2019 & YFCC100M~\cite{thomee2016yfcc100m} \\ 
VidOR~\cite{shang2019vidor}            & VidSGG & 10,000  &  82  & 30s &  3rd   & 35  & 25 & Bounding Box  & - & 2020 & Charades~\cite{sigurdsson2016hollywood} \\ 
\midrule
Cityscapes-VPS~\cite{kim2020video}   & VPS & 500  &  -  & - &  vehicle   & 19  & - & Panoptic Seg.  & 3K & 2020 & - \\ 
KITTI-STEP~\cite{weber2021step}       & VPS & 50  &  -  & -  &  vehicle   & 19  & - & Panoptic Seg.  & 18K & 2021 & - \\ 
VIP-Seg~\cite{miao2022large}          & VPS & 3,536  &  5  & 5s &  3rd   & 124  & - & Panoptic Seg.  & 85K & 2022 & - \\ 
\midrule
Ego4D-STA~\cite{grauman2022ego4d}        & STA & 1,498  &  111  & 264s &  ego   & -  & - & Bounding Box  & - & 2022 & - \\ 
VISOR~\cite{darkhalil2022epic}   & VOS & 179  &  36  & 720s &  ego   & 257  & 2 & Semantic Seg.  & 51K & 2022 & EPIC-KITCHENS~\cite{damen2022epic} \\ 
\midrule
\textbf{PVSG}    & PVSG & 400  &  9  & 77s &  3rd + ego   & 126  & 57 & Panoptic Seg.  & 150K & 2023 & VidOR + Ego4D + EPIC-KITCHENS \\ 
\bottomrule
\end{tabular}
}
\vspace{-16pt}
\end{table*}

In the last several years, scene graph generation has received increasing attention from the computer vision community~\cite{johnson2015image,xu2017scene,yang2018graph,ji2020actiongenome,li2022embodied,xu2022meta,yang2022panoptic}. 
Unlike object-centric labels like ``person'' or ``bike'', or the precise bounding boxes typical in object detection, scene graphs offer a richer representation of images by capturing both objects and their pairwise relationships and/or interactions, such as ``a person riding a bike''. A notable trend in this field is the evolution from static, image-based scene graphs to dynamic, video-level scene graphs~\cite{xu2022meta,teng2021target,chen2022adaptive}, marking a significant advancement towards more comprehensive visual perception systems.

While videos undoubtedly provide richer information than individual images due to the additional temporal dimension, which greatly aids in understanding temporal events~\cite{herath2017going}, reasoning~\cite{zhou2018temporal}, and identifying causality~\cite{fire2015learning}, current video scene graph representations, primarily based on bounding boxes, still fall short of replicating human visual perception. This gap can be attributed to their lack of \textit{granularity}, a limitation that can be overcome by integrating \textit{panoptic segmentation masks}.
This is echoed by the evolutionary trajectory in visual perception research, progressing from image-level labels (i.e., classification) to spatial locations (i.e., object detection), and finally to more fine-grained, pixel-wise masks (i.e., panoptic segmentation~\cite{kirillov2019panoptic}).

In this paper, we take scene graphs to the next level by proposing \textit{panoptic video scene graph generation (PVSG)}, a new problem that requires each node in video scene graphs to be grounded by a pixel-level segmentation mask. Panoptic video scene graphs address a critical limitation in bounding box-based video scene graphs: comprehensively covering both ``things'' and ``stuff'' classes (i.e., amorphous regions such as water, grass, etc.), with the latter being essential for contextual understanding yet challenging to localize with bounding boxes.
For instance, when applying PVSG to the video in Figure~\ref{fig:teaser}, relations like ``adult-1 standing on the ground'' and ``adult-2 standing in water'' become evident, which are typically overlooked in bounding box-based scene graphs. Furthermore, existing bounding box-based annotations~\cite{ji2020actiongenome} often overlook small yet significant details, for example, ``candles on cake".

To help the community progress in this new area, we contribute a carefully annotated PVSG dataset, comprising 400 videos (289 third-person and 111 egocentric) with an average duration of 76.5 seconds each. This dataset encompasses around 150K frames, all annotated with detailed panoptic segmentation and temporal scene graphs, covering 126 object classes and 57 relation classes. A comprehensive comparison of our PVSG dataset with related datasets is shown in Table~\ref{tab:dataset}.

Our solution to the PVSG challenge involves a two-stage framework. The first stage generates a set of feature tubes for each mask-based instance tracklet, while the second stage constructs video-level scene graphs based on these tubes. We explore two options for the first stage: 1) combining an image-level panoptic segmentation model with a tracking module, and 2) employing an end-to-end video panoptic segmentation model. For the scene graph generation stage, we present four distinct implementations, encompassing both convolutional and Transformer-based methods.

In summary, we make the following contributions to the scene graph community:
\begin{enumerate}
    \item \textbf{A New Problem}: We identify several issues associated with current research in video scene graph generation and propose a new problem, which combines video scene graph generation with panoptic segmentation for holistic video understanding.
    \item \textbf{A New Dataset}: A high-quality dataset with fine, temporal scene graph annotations and panoptic segmentation masks is proposed to advance the area of PVSG.
    \item \textbf{New Methods and Benchmarking}: We propose a two-stage framework to address the PVSG problem and benchmark a variety of design ideas, providing valuable insights for future research in this domain.
\end{enumerate}

We have released two key codebases in this project: PVSGAnnotation\footnote{\url{https://github.com/LilyDaytoy/PVSGAnnotation}} for the video panoptic segmentation annotation pipeline, and OpenPVSG\footnote{\url{https://github.com/LilyDaytoy/OpenPVSG}} to benchmark PVSG methods, both aimed at aiding the community in further research.

%% file: section/2_related_work.tex
\section{Related Work}
\label{sec:related_work}

\begin{figure*}[t]
    \centering
    \includegraphics[width=\linewidth]{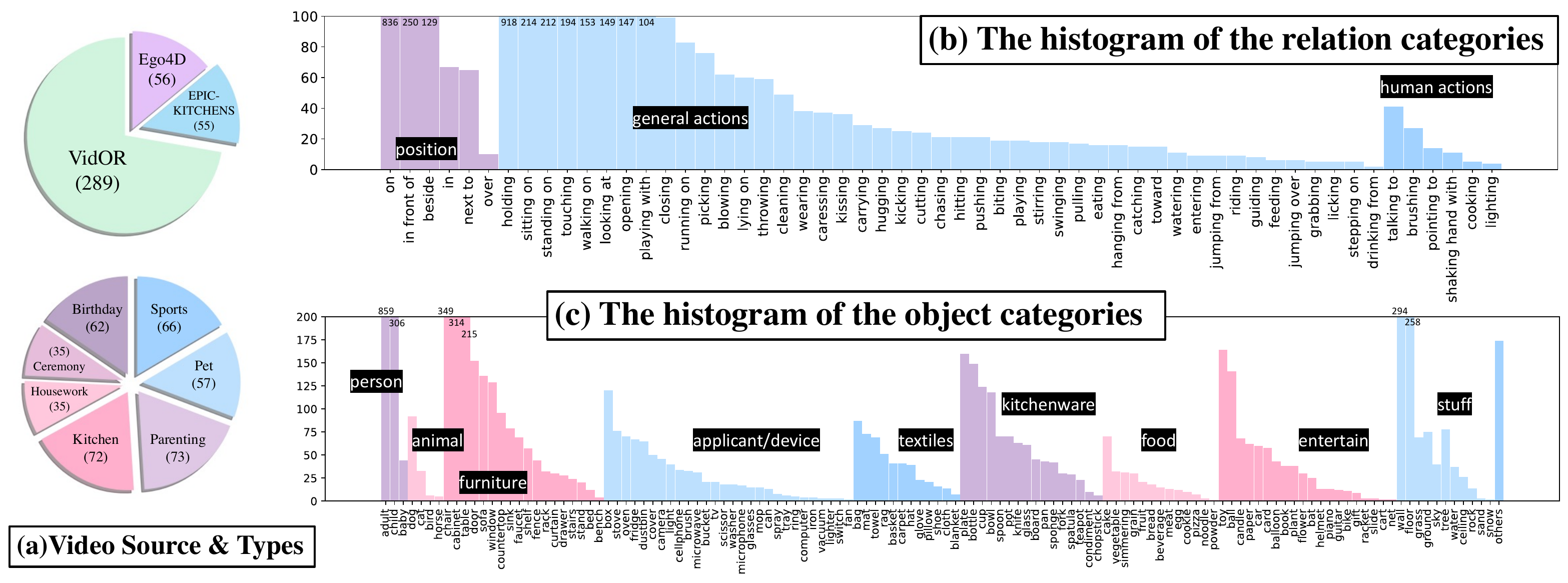}
    \vspace{-20pt}
    \caption{\textbf{The PVSG dataset statistics.} 
    The PVSG dataset contains 400 third-person and ego-centric videos from diverse environments, as shown in (a). The statistics of object classes and relation classes are shown in (b) and (c).
    }
    \vspace{-10pt}
    \label{fig:pvsg_stat}
\end{figure*}

\paragraph{Scene Graph Generation}
Given an image, the scene graph generation (SGG) task expects the model to output a scene graph representation, where nodes represent objects and edges represent relations between objects. To localize object instances, the nodes should be grounded by the bounding boxes~\cite{xu2017scene}. Classic scene graph generation methods have been dominated by the two-stage pipeline that consists of object detection and pairwise predicate estimation~\cite{tang2018vctree, xu2017scene, zellers2018neural, suhail2021energybased, zhong2021learning}. Recent works on one-stage methods~\cite{li2022sgtr, yang2022panoptic, cong2022reltr} provide simpler models that output semantically diverse relation predictions. Though the prevalent SGG benchmark Visual Genome~\cite{vg17ijcv} provides rich annotations, it suffers from numerous “noisy” ground-truth predicate labels, e.g., some unannotated negative samples are not absolutely background. NICE~\cite{li2022devil} reformulates SGG as a noisy label learning problem. They re-assign pseudo labels to detect noisy negative samples. Instead of exploiting the noisy SGG datasets, recently a new task of panoptic scene graph generation (PSG)~\cite{yang2022panoptic} has been proposed with a refined PSG dataset, based on panoptic segmentation annotations to identify foreground and background concretely. Our work extends PSG to the video level by predicting spatial-temporal relations.

\paragraph{Video Scene Graph Generation}
Shang~\etal~\cite{shang2017video} first proposes Video Scene Graph Generation (VidSGG) and released the ImageNet-VidVRD dataset. They generate object tracklet proposals and short-term relations on overlapping segments. Subsequently, they greedily associate these relation triplets into video level. Several works follow the track-to-detect paradigm with spatio-temporal graph and graph convolutional neural networks~\cite{qian2019video, liu2020beyond}, or multiple hypothesis association~\cite{su2020video}. MVSGG~\cite{xu2022meta} investigates the spatio-temporal conditional bias problem in VidSGG. They perform a meta-training and testing process, constructing the data distribution of each query set w.r.t. the conditional biases. TRACE~\cite{teng2021target} decouples the context modeling for relation prediction from the complicated low-level entity tracking. \cite{chen2022adaptive} raises the issue of domain shift between image and video scene graphs. They exploit external commonsense knowledge to infer the unseen dynamic relationship and employ hierarchical adversarial learning to adapt from image to video data distributions. Embodied Semantic SGG~\cite{li2022embodied} exploits the embodiment of the intelligent agent to autonomously generate an appropriate path by reinforcement learning~\cite{dong2021baconian} to explore an environment.
 
\paragraph{Video Panoptic Segmentation} Video Panoptic Segmentation (VPS)~\cite{STEP,kim2020vps,miao2022large} unifies both Video Semantic Segmentation~\cite{cordts2016cityscapes} and Video Instance Segmentation~\cite{vis_dataset} in one framework. It extends panoptic segmentation into video by making instance IDs across frames consistent. VPSNet~\cite{kim2020vps} first extends cityscapes sequences~\cite{cordts2016cityscapes} and builds a VPS dataset for driving scene, along with a new metric named Video Panoptic Quality (VPQ). STEP dataset~\cite{STEP} proposes another metric named Segmentation and Tracking Quality (STQ) that decouples the segmentation and tracking error. VIP-Seg~\cite{miao2022large} proposes a large-scale VPS dataset which contains various scenes. Several works~\cite{kim2020vps,woo2021learning_associate_vps,yuan2021polyphonicformer} are proposed to solve VPS task respectively. VIP-Deeplab~\cite{ViPDeepLab} extends the Panoptic-Deeplab~\cite{cheng2020panoptic} with the next frame center map prediction. Video K-Net~\cite{li2022videoknet} unifies the VPS pipeline via kernel online tracking and linking. TubeFormer~\cite{kim2022tubeformer} process tube-frames with temporal attention. Compared with previous VPS datasets, our PVSG dataset contains extremely long videos, which bring new challenges for VPS tasks. Moreover, our work goes beyond VPS tasks by also considering relations across a video.


%% file: section/3_problem.tex
\section{The PVSG Problem}
\label{sec:problem}
The goal of the PVSG problem is to describe a given video with a dynamic scene graph, with each node associated with an object and each edge associated with a relation in the temporal space. 
Formally, the input of the PVSG model is a video clip $\mathbf{V}\in \mathbb{R}^{T\times H\times W\times 3}$, where $T$ denotes the number of frames, and the frame size $H\times W$ should be consistent across the video. The output is a dynamic scene graph $\mathbf{G}$. The PVSG task can be formulated as follows,
\begin{equation}
\label{E:sgg_def}
\operatorname{Pr}\left(\mathbf{G} \mid \mathbf{V}\right)=\operatorname{Pr}\left(\mathbf{M},\mathbf{O},\mathbf{R} \mid \mathbf{V}\right).
\end{equation}
More specifically, $\mathbf{G}$ comprises the binary mask tubes $\mathbf{M}=\left\{\mathbf{m}_{1}, \ldots, \mathbf{m}_{n}\right\}$ and object labels $\mathbf{O}=\left\{o_{1}, \ldots, o_{n}\right\}$ that correspond to each of the $n$ objects in the video, and their relations in the set $\mathbf{R}=\left\{r_{1}, \ldots, r_{l}\right\}$. 
For object $i$, the mask tube $\mathbf{m}_{i} \in\{0,1\}^{T \times H \times W}$ collects all its tracked masks in each frame, and its object category should be $o_i\in\mathbb{C}^O$.
For all objects in a frame $t$, the masks do not overlap, i.e., $\sum_{i=1}^{n} \mathbf{m}^{t}_{i} \leq \mathbf{1}^{H \times W}$.
The relation $r_i\in\mathbb{C}^R$ associates a subject and an object with a predicate class and a time period. $\mathbb{C}^O$ and $\mathbb{C}^R$ means the object and predicate classes.

\paragraph{Metric}
In practice, the output of the PVSG task is to predict a set of triplets to describe the input video. Take a triplet as an example, which contains a relation $r_i$ from $t_1$ to $t_2$, associates the subject with its class category $o_s$ and mask tube $\mathbf{m}_{s}^{(t_1,t_2)}$, and an object with $o_s$ and $\mathbf{m}_{o}^{(t_1,t_2)}$. 
$\mathbf{m}^{(t_1,t_2)}$ denotes the mask tube $\mathbf{m}$ span across the period of $t_1$ to $t_2$.

\begin{figure}[t]
    \centering
    \includegraphics[width=\linewidth]{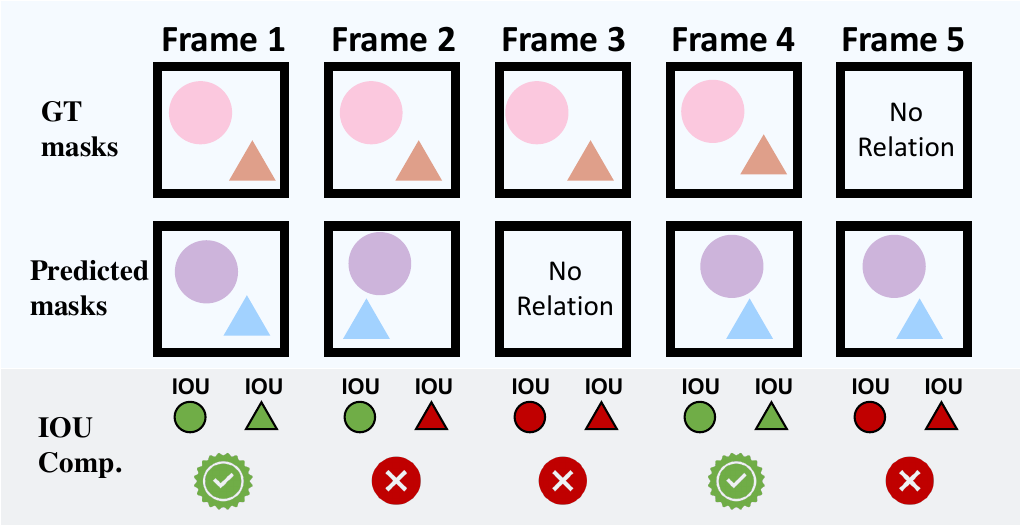}
    \vspace{-20pt}
    \caption{\textbf{Illustration of the PVSG Metric.} Assuming the classification of the triplet is correct, to further match the ground truth (GT) frame-wise, the predicted mask pair must have both subject and object masks with a mask IOU above 0.5. In this case, only Frames 1 and 4 satisfy this condition, yielding an intersection count of 2 and a union count of 5. Thus, the volume IOU is calculated as 0.4. As this value falls short of the 0.5 threshold, it is not considered a successful recall.
    }
    \vspace{-20pt}
    \label{fig:metric}
\end{figure}

To evaluate the PVSG task, we follow the classic SGG and VidSGG paper and use the metrics of the R@K and mR@K, which calculates the triplet recall and mean recall given the top K triplets from the PVSG model.
A successful recall of a ground-truth triplet ($\hat{o}_s$, $\mathbf{\hat{m}}^{(\hat{t}_1,\hat{t}_2)}_{s}$, $\hat{o}_s$, $\hat{\mathbf{m}}^{(\hat{t}_1,\hat{t}_2)}_{o}$, $\hat{r}_i^{(\hat{t}_1,\hat{t}_2)}$) should meet the following criteria: 1) the correct category labels of the subject, object, and predicate; 2) the predicted mask tubes ($\mathbf{m}_{s}^{(t_1,t_2)}$, $\mathbf{m}_{o}^{(t_1,t_2)}$) and the ground-truth tubes ($\mathbf{\hat{m}}_{s}^{(\hat{t}_1,\hat{t}_2)}$, $\mathbf{\hat{m}}_{o}^{(\hat{t}_1,\hat{t}_2)}$) should have volume IOU over 0.5. More specifically, we compute the time IOU between the ground-truth $(t_1,t_2)$ and $(\hat{t}_1,\hat{t}_2)$, and the frame $t$ is considered as intersection only when both $\mathbf{m}_{s}^t$ and $\mathbf{m}_{o}^t$ have mask IOU over 0.5 compared to their ground-truth counterpart. Figure~\ref{fig:metric} shows how volume IOU calculates. Following the scene graph generation conventions, we adopt a 0.5 threshold for time IOU as the standard for considering a triplet recalled. Additionally, in the experiment, we also report results with the threshold of 0.1, a lower standard relaxes the criteria for time span prediction.

Please notice the nuance of the PVSG metrics compared with VidSGG metrics for VidOR~\cite{shang2019vidor}. For a case where a child stops and goes several times in a video, different from VidOR which considers several ``child-1 walking on ground'' triplets, our PVSG metrics only consider the triplet once, but with a scattered time span. This small change avoids some relations dominating the metrics by repeating.

\begin{figure*}
    \centering
    \includegraphics[width=\linewidth]{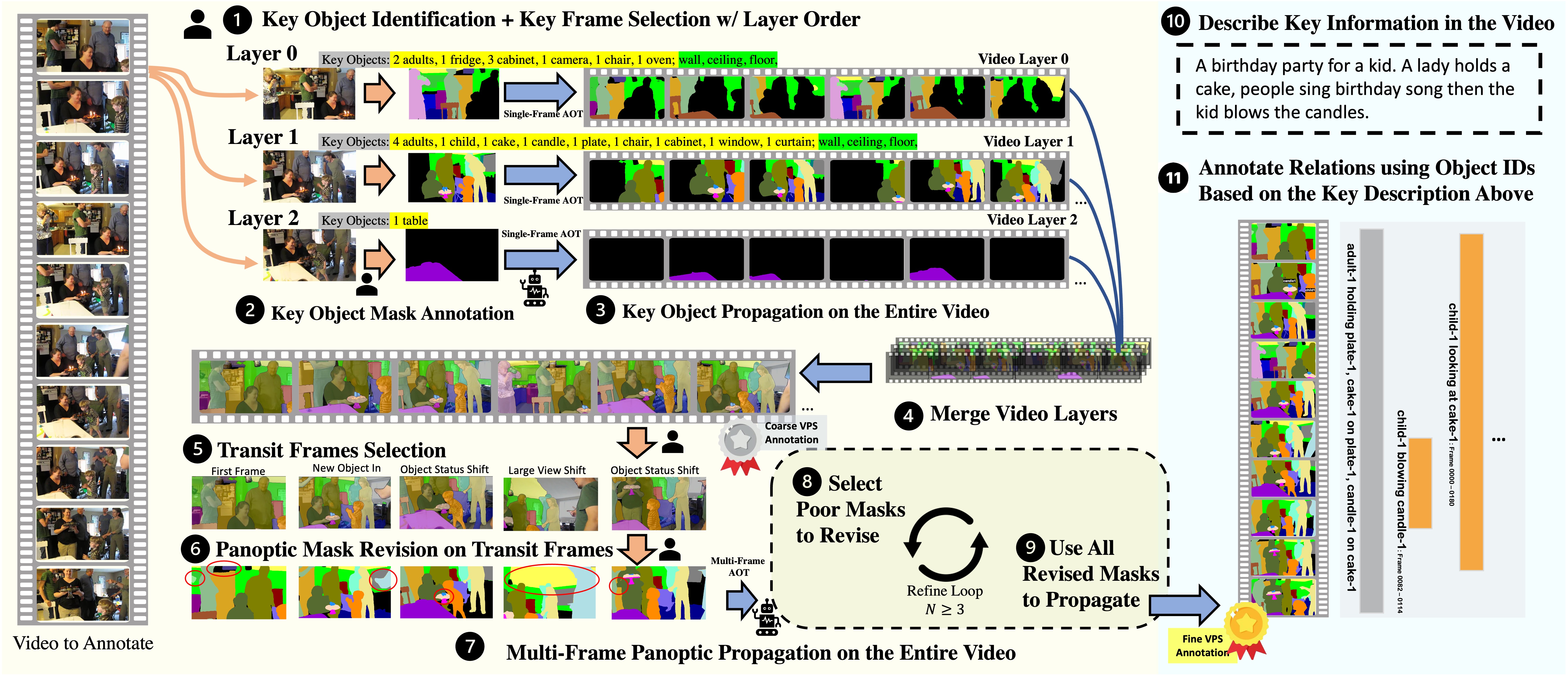}
    \vspace{-22pt}
    \caption{\textbf{PVSG Dataset Annotation Pipeline.} 
    The construction of the PVSG dataset can be divided into VPS annotation and relation annotation. For VPS annotation, we select a few key frames and use an off-the-shelf video object segmentation (VOS) model AOT~\cite{yang2021aot} to propagate the annotated objects to the whole video, and then perform frame-level mask fusion using the predefined layer order to obtain a coarse VPS annotation for further revision.
    The relations are annotated based on the description of the key information in the video.}
    \label{fig:pipeline}
    \vspace{-12pt}
\end{figure*}

%% file: section/4_dataset.tex
\section{The PVSG Dataset}
\label{sec:pvsg_dataset}

In this section, we first summarize the existing VidSGG datasets and highlight their problems. Then, we introduce the overview and statistics of our PVSG dataset and its annotation pipeline.

\subsection{Connecting Existing Datasets to PVSG}
To select candidate video clips for the PVSG dataset, a go-to option is to borrow the videos from other VidSGG datasets. 
Table~\ref{tab:dataset} lists three classic VidSGG datasets chronologically. 
While the limited size of their first VidSGG dataset, ImageNet-VidVRD~\cite{shang2017video}, Shang~\etal collects 10K videos from the user-uploaded dataset YFCC100M~\cite{thomee2016yfcc100m} and generate a large-scale VIDOR dataset~\cite{shang2019vidor},
with dense object and relation annotation. Ji~\etal also introduces a large-scale dataset Action Genome (AG) based on a diverse, crowd-sourcing Charades dataset~\cite{sigurdsson2016hollywood}.
While Charades provides a novel solution to gather large-scale, less-biased video datasets by asking people to act based on the generated script, the curated scripts usually produce random action series, such as a man rushing out of the room and running back for no reason. Also, the performance traces turn out to be heavy in the dataset. These shortcomings limit the potential of the community to explore contextual logic and reasoning in videos.

Alternative video datasets that lean toward logic reasoning and video scene understanding are instruction datasets or movie datasets. However, these datasets are either full of close-up shots (e.g., Something-Something~\cite{goyal2017something}, Howto100M~\cite{miech19howto100m}) or cut shots (e.g., MOMA~\cite{luo2021moma}, HC-STVG~\cite{tang2021human}).
In fact, humans rely on unpolished videos to form an essential understanding of the world. In this sense, we find that the unedited, natural, and diverse VidOR~\cite{shang2019vidor} videos are a good candidate for learning the visual essence as well as keeping the potential of contextual logic exploration.
While the videos presented above showcase a third-person perspective, egocentric videos have gained popularity due to their practicality in autonomous driving~\cite{yao2019egocentric}, robotic decision-making~\cite{zhang2013video}, and in the metaverse~\cite{ooi2022sense}. In particular, a subset of the Ego4D dataset~\cite{grauman2022ego4d} is suitable for exploring logical relationships and modeling, as it supports short and long-term action anticipation tasks. Additionally, the Epic-Kitchens~\cite{damen2018scaling} dataset is focused on the kitchen scenario and offers rich action data. Its subset, the VISOR dataset, includes video object segmentation (VOS) annotation, which partially matches the PVSG scope, though its relations are not yet annotated.

Another dataset category that is closely related to the PVSG problem is the video panoptic segmentation (VPS) datasets. Popular VPS datasets include Cityscapes-VPS~\cite{kim2020video} and KITTI-STEP~\cite{weber2021step}. However, the relations in the self-driving scenarios are limited, which is not suitable for the PVSG task. 
Although the recent VIP-Seg~\cite{miao2022large} provides a more diverse VPS dataset, each video only lasts around 5 seconds, which also lacks temporal relations.

With all the rationale above, we eventually decide to combine three video sources to the PVSG dataset, which are VidOR, Ego4D-STA, and Epic-Kitchens-100 (including some videos from VISOR).

\subsection{Dataset Statistics}
Figure~\ref{fig:pvsg_stat} displays the statistics of the PVSG dataset, which consists of 400 videos, including 289 third-person videos from VidOR and 111 egocentric videos from Epic-Kitchens and Ego4D. Among the videos, 62 videos feature birthday celebrations, while 35 videos center around ceremonies, providing rich content for contextual logic and reasoning. Furthermore, the dataset includes numerous videos related to sports and pets, featuring complex and diverse actions and interactions between objects. Figure~\ref{fig:pvsg_stat} (c) shows the object count (including stuff) in the PVSG dataset.

\subsection{Dataset Construction Pipeline}
Creating the PVSG dataset is never a trivial task considering that both video panoptic segmentation and relation annotations are required. This section describes how the PVSG dataset is collected and annotated.

\paragraph{Step 1: Video Clip Selection}
To get rid of the drawbacks of the current datasets (i.e., the unnatural videos in AG~\cite{ji2020actiongenome} without logical script, and the static and short videos from the VPS datasets), we carefully select around 300 long, daily, unedited videos with a logical storyline.
In addition, to encourage the VidSGG models to be practical on egocentric videos, we also select around 100 videos from Epic-Kitchens and Ego4D with the same criteria.
Videos with too many small and trivial objects are also discarded for VPS annotation purposes.
We hope the selected videos could greatly encourage the exploration of video recognition, understanding, and reasoning.

\begin{figure*}
    \centering
    \includegraphics[width=\linewidth]{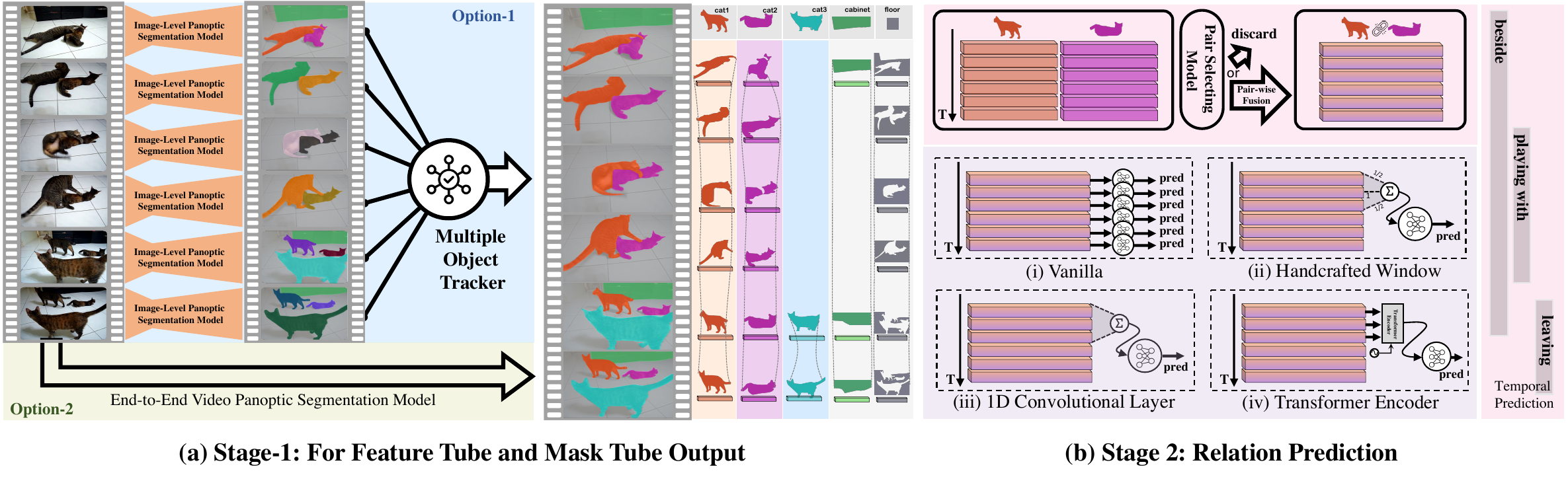} 
    \vspace{-0.7cm}
    \caption{
    \textbf{The two-stage framework to solve the PVSG task.} 
    The goal of the first stage is to obtain the video panoptic segmentation mask for each object, as well as its corresponding video-length feature tube. Two options are provided to achieve the goal.
    The second stage predicts pairwise relations based on all the feature tubes from the first stage. Four options are provided for a comprehensive comparison.
    }
    \vspace{-10pt}
    \label{fig:method}
\end{figure*}

\paragraph{Step 2: VPS Annotation}
Notice that the PVSG videos have more than 300 frames on average and 150K in total, it is impossible to annotate panoptic segmentation for each frame. After iterations and improvements, we finalize a human-machine collaborative VPS annotation pipeline, depicted in Figure~\ref{fig:pipeline}. In a nutshell, we largely rely on an off-the-shelf VOS model called AOT~\cite{yang2021aot} for the human-machine interactive annotation process.

\noindent\textbf{Coarse VPS Annotation:}
With a few well-annotated object masks in the first frame, the AOT~\cite{yang2021aot} is able to propagate the masks to later frames. With this strong automatic tool, we design a pipeline to obtain coarse VPS annotation. 
For the example video in Figure~\ref{fig:pipeline} (actions 1-3), we first identify several key objects to annotate and also identify key frames where the selected objects have a clear and whole appearance. 
To identify key objects, our annotators need to select all objects and backgrounds to address ``panoptic'', except those messy and unrelated ones. 
After annotating these key objects on their corresponding frames, we use AOT based on the frames to propagate the mask, both forward and backward. Thus, each frame will yield a whole mask video. To merge those mask videos into one, the layer order should be considered beforehand, i.e., the objects from which layer should be put in front. In fact, the decision of the layer order is made with keyframe selection.

\noindent\textbf{Fine VPS Annotation:}
Based on the coarse VPS annotation, we conduct several rounds (more than 5) of the human-machine interactive revision process to obtain the final annotation. We rely on the multi-frame panoptic segmentation propagation mode of the AOT algorithm~\cite{yang2021aot}, which interpolates the entire video mask based on several frames with the entire panoptic segmentation. The quality of interpolation increases with more intermediate frames. To accelerate the revision process, we revise the transit frames first, as shown in action 5 in Figure~\ref{fig:pipeline}. Typical examples of poor masks include incorrect tracking masks and boundaries that deviate significantly from the object.

\paragraph{Step 3: Relation Annotation}
We annotate temporal relations based on the VPS annotation, with object ID prepared. To guarantee the significance of the relation, we ask annotators to describe the video with several sentences and annotate relations accordingly. The relations they use are strictly within our dictionary, but we also enlarge the dictionary when necessary. Similar to the PSG dataset~\cite{yang2022panoptic}, we ask the annotators to use the most unambiguous predicate possible, i.e., ``sitting on'' rather than ``on''.

%% file: section/5_baselines.tex
\section{Methodology}
\label{sec:method}
In this section, we introduce the two-stage pipeline to address the PVSG problem. We provide two options for the first stage and four options for the second stage.

\begin{table*}[t]
\caption{\textbf{Comparison between all two-stage PVSG baselines.} We provide two options for the first stage and four options for the second stage, as described in Section 3. The results show that using the basic image-based method in the first stage with the transformer encoder in the second stage can achieve optimal recall.
}
\vspace{-10pt}
\label{tab:pvsg}
\small
\centering
\setlength\tabcolsep{10pt}
\resizebox{\linewidth}{!}{
\begin{tabular}{@{\hskip 0.3in}c@{\hskip 0.5in}l@{\hskip 0.5in}ccc@{\hskip 0.5in}ccc@{\hskip 0.3in}}
\toprule
\multicolumn{2}{@{\hskip 1.0in}l}{Method} & \multicolumn{3}{c}{thre $=0.5$} & \multicolumn{3}{c}{thre $=0.1$} \\ \midrule
Stage-1 & Stage-2 & R/mR@20 & R/mR@50 & R/mR@100 & R/mR@20 & R/mR@50 & R/mR@100 \\ \midrule
\multicolumn{1}{c}{\multirow{4}{*}{IPS+T~\cite{cheng2021mask2former, wangUnitrack}}} 
 & Vanilla & 3.04 / 1.35&4.61 / 2.94&5.56 / 3.33&8.28 / 5.68&14.47 / 9.92&18.24 / 11.84 \\
 & Handcrafted Window & 2.52 / 1.72&3.77 / 2.36&4.72 / 2.79&8.07 / 5.61&13.42 / 8.27&16.46 / 10.11\\
 & 1D Convolution & \textbf{3.88} / 2.55&5.24 / 3.29&\textbf{6.71} / \textbf{5.36}&\textbf{10.06} / \textbf{8.98}&\textbf{14.99} / \textbf{12.21}&\textbf{18.13} / \textbf{15.47} \\
 & Transformer Encoder & \textbf{3.88} / \textbf{2.81}&\textbf{5.66} / \textbf{4.12}&6.18 / 4.44&9.01 / 6.69&14.88 / 11.28&17.51 / 13.20 \\
\midrule
\multicolumn{1}{c}{\multirow{4}{*}{VPS~\cite{li2022videoknet, cheng2021mask2former}}} & Vanilla & 0.21 / 0.10 & 0.21 / 0.10 & 0.31 / 0.18& 6.29 / 3.04& 9.64 / 6.74& 12.89 / 9.60 \\
\multicolumn{1}{c}{} & Handcrafted Window & \textbf{0.42} / 0.13& 0.52 / 0.50& 0.94 / \textbf{0.92}& 5.24 / 2.84& 7.65 / 7.14& 9.64 / 8.22 \\
\multicolumn{1}{c}{} & 1D Convolution & \textbf{0.42} / 0.25& 0.63 / 0.67& 0.63 / 0.67& \textbf{8.07} / \textbf{7.84}& \textbf{11.01} / \textbf{9.78}& \textbf{12.89} / \textbf{10.77} \\
\multicolumn{1}{c}{} & Transformer Encoder & \textbf{0.42} / \textbf{0.61}& \textbf{0.73} / \textbf{0.76}& \textbf{1.05} / \textbf{0.92}& 6.50 / 5.75& 9.64 / 8.25& 12.26 / 9.51 \\
\bottomrule 
\end{tabular}
}
\vspace{-10pt}
\end{table*}

\subsection{Stage One: Video Panoptic Segmentation}
Given a video clip input $ \mathbf{V} \in \mathbb{R}^{T\times H\times {W}\times 3}$, the goal of VPS is to segment and track each pixel in a non-overlap manner.
Specifically, the model predicts a set of video clips $\{y_i\}_{i=1}^N = \{(\mathbf{m}_i, p_i(c))\}_{i=1}^N$, where $\mathbf{m}_i \in \{0,1\}^{T\times H\times W}$ denotes the tracked video mask, and $p_i(c)$ denotes the probability of assigning class $c$ to a clip $\mathbf{m}_i$. $N$ is the number of entities, which includes thing classes and stuff classes.
 
We present two strong baselines for the first stage of VPS processing. In particular, we adopt the state-of-the-art image segmentation baseline~\cite{cheng2021mask2former} with an extra tracker and the improved video panoptic segmentation
method~\cite{li2022videoknet}. For the former, it processes the video frames individually. For the latter, it processes the video frames across the temporal dimension, with a nearby frame as the reference frame.

\paragraph{IPS+T: Image Panoptic Segmentation With Tracker}  We adopt strong Mask2Former~\cite{cheng2021mask2former} as our baseline method since it is a mask-based transformer architecture. It contains a transformer encoder-decoder architecture with a set of object queries, where the object queries interact with encoder features via masked cross-attention. Given an image $\mathbf{I}$, during the inference, the Mask2Former directly outputs a set of object queries $\{q_{i}\}, i=1,\dots, N$, where each object query $q_{i}$ represent one entity.
Then, two different multiple-layer perceptrons (MLPs) project the queries into two embeddings for mask classification and mask prediction, respectively. During training, each object query is matched to ground truth masks via masked-based bipartite matching. 

We first fine-tune the Mask2Former on our dataset. Then, we test the model with an extra tracker~\cite{wangUnitrack}. In particular, we first obtain panoptic segmentation results of each frame. Then we link each frame via using UniTrack~\cite{wangUnitrack} for tracking to obtain the final $N$ tracked video cubes for each clip. Therefore, a query tube is obtained. For the object $i$ at the $t$-th frame, the query is noted as $q_i^t$. We use $\mathbb{Q}^{(t_1, t_2)}_i$ to denote the set of queries $\{q^t_i\}_{t=t_1}^{t_2}$, and $\mathbb{Q}_i$ denotes the query tube in the entire video.

\paragraph{VPS: Video Panoptic Segmentation Baseline} For video baselines, we modify the previous state-of-the-art method Video K-Net~\cite{li2022videoknet} into Mask2Former framework. We first replace the backbone and neck in Video K-Net~\cite{li2022videoknet} with the Mask2Former feature extractor. Then we use the temporal contrastive loss to perform directly on the output queries from the last layer of the decoder. In particular, given two frames, we first obtained the object queries from both frames and then we sent them into an embedding layer (a shared MLP) to obtain association embeddings.
We adopt the same tracking loss used in ~\cite{li2022videoknet} to supervise the association embeddings. The embeddings are close if they are the same object, otherwise, they are pulled away. 

During the training, the two nearby frames are sent to the model to learn the association embedding. During the inference, the learned association embeddings are used to perform instance-wised tracking cues to match each thing masks frame by frame in an online manner. Compared with the image baseline, our video baseline considers the temporal learned embedding. After this step, we obtain $N$ tracked video cubes for each clip. For both baselines, we also dump the corresponding object queries for further processing.

\subsection{Stage Two: Relation Classification}
The object query (feature) tubes $\{Q_i\}_{i=1}^N$ serve as a link between the first and second stages. As shown in Figure~\ref{fig:method} (b), these tubes are initially formed into query pairs. For efficient training of the relation model, these pairs are then matched with their corresponding ground-truth pairs based on mask IOU values, with non-matching pairs being discarded. This selective process assigns relation labels to certain predicted query pairs during specific time spans.

\noindent\textbf{Pair Selection} \quad It is important to note the difference in pairing selection between the training and inference phases. During training, pairs are easily selected based on their match with the ground truth. However, at inference, pairing all possible combinations would yield $N \times (N-1)$ pairs, an impractically large number. To address this challenge, we have developed a compact, trainable pairing component~\cite{wang2023pair}. This component leverages a transformer encoder to cross-attend to all other object features within each frame, thus gathering global information. Subsequently, it uses max pooling to condense the query tube $\{Q^t_i\}^T_{t=t_0}$ into a single token for each object. This process allows for the calculation of pair-wise similarities and the construction of a sparse pairing matrix, which is optimized towards the ground truth pairing matrix using a multi-label loss~\cite{su2022zlpr}.

Next, we introduce four operation options to predict the relations of each feature pair.

\noindent\textbf{Vanilla: Fully-Connected Layers}\quad
Begin with the most basic version, the pairwise feature fusion is followed by three straightforward fully-connected layers on the fused features. Since some objects may have several interactions occurring simultaneously, we define the issue as a multi-label classification job with binary cross-entropy loss.

\noindent\textbf{Handcrafted Filter}\quad
To further consider the temporal information, we design a simple kernel to gather the information from the context in nearby frames. By default, the handcrafted filter is a simple vector of $[\frac{1}{4}, \frac{1}{2}, 1, \frac{1}{2}, \frac{1}{4}]$ with a window size of $5$. The filter is also required in inference.

\noindent\textbf{1D-Convolutional Layer}\quad
To improve the handcrafted filter, we also utilize a learnable 1D-Convolutional layer to capture temporal information. The kernel sizes are set to $5$.

\noindent\textbf{Transformer Encoder}\quad
A transformer encoder~\cite{vaswani2017attention} is naturally suitable for encoding the temporal data. We utilize a transformer block with positional embeddings in the entire fused query feature to capture temporal information via cross-attention between frames.

\begin{figure*}[t]
    \centering
    \includegraphics[width=\linewidth]{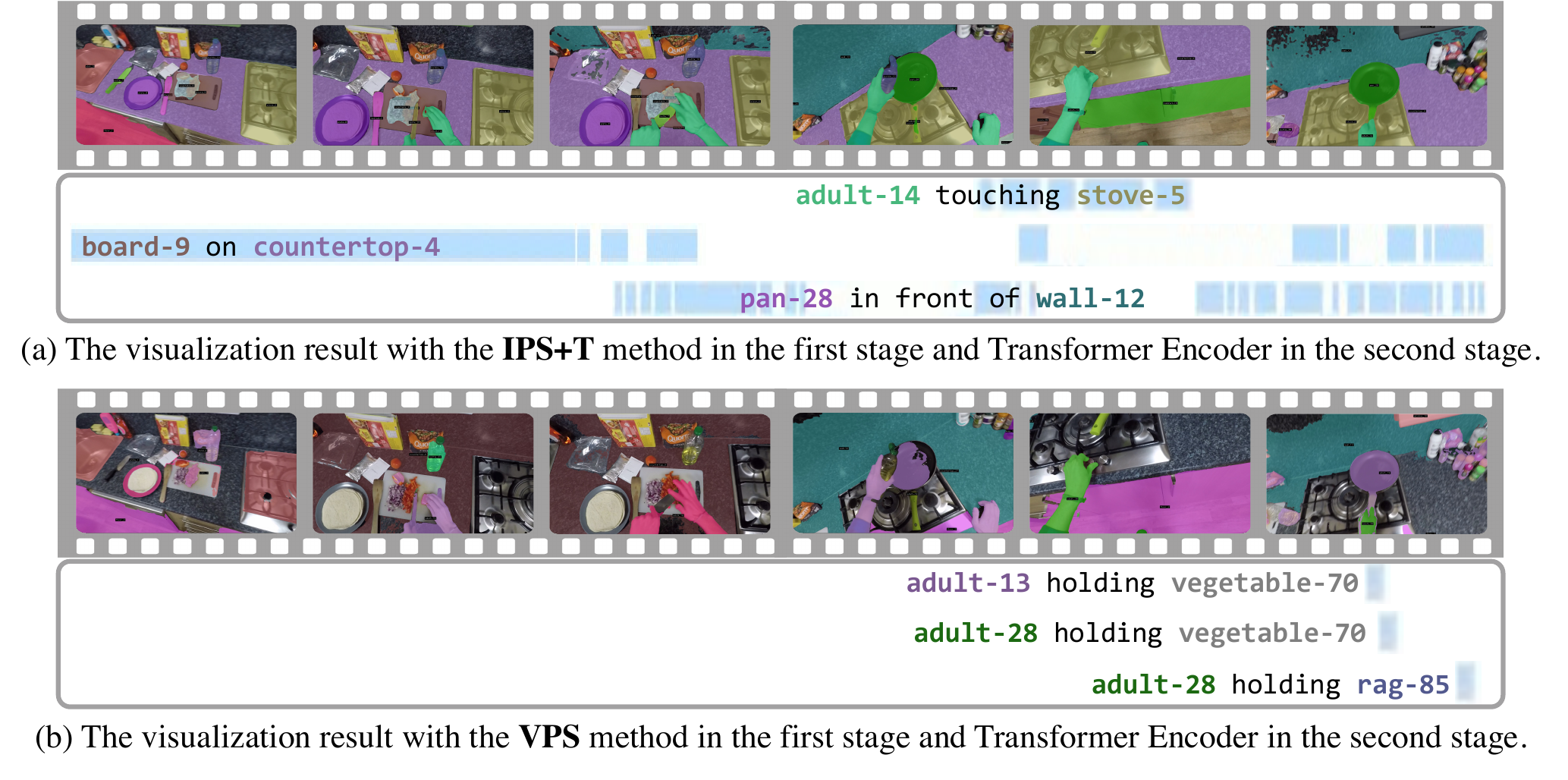} 
    \vspace{-25pt}
    \caption{\textbf{The visualization of the top 3 triplets generated by PVSG models.} The result shows that the IPS+T method is able to predict a better-quality video panoptic mask. The VPS baseline is shown unable to perform well on tracking (e.g., the tracking of the adult switched in the later frames), which leads to its low performance in the PVSG task. Check project page for more visual results.}
    \label{fig:exp_vis}
\end{figure*}

%% file: section/6_results.tex
\section{Experiments}
\label{sec:exp}
In this section, we show the experimental results for the PVSG dataset. We split the dataset with 338 videos for training and 62 videos for testing\footnote{Check the annotated videos in each split \href{https://entuedu-my.sharepoint.com/:f:/g/personal/jingkang001_e_ntu_edu_sg/EgvpTfCTMudLpxw-h0_BVdcBAHacUaAQD-u9OvkUlpaDBg?e=4462g0}{here}.}. 
For both IPS+T and VPS, we adopt Mask2Former~\cite{cheng2021mask2former} upon the ResNet-50~\cite{he2016deep} backbone with 8 training epochs, both take about 48 hours on 4 V-100 GPUs. The training epoch of the second stage is set to 100, which takes about 8 hours on one V-100.

\paragraph{Better Temporal Modeling Boosts Relations Prediction.} We first take a look at the second stage. The transformer encoder obtains the optimal results regardless of the first-stage options, underscoring its proficiency in synthesizing temporal information. Moreover, the 1D convolutional approach outperforms the handcrafted window method, suggesting that incorporating learnable parameters in the second stage can be beneficial. Notably, even the most elementary vanilla method registers some recall considering the harsh recall criteria described in Section~\ref{sec:problem}. This indicates that with a decent model in the first stage, the PVSG task is indeed approachable.

\paragraph{VPS Models Lag Behind IPS+T.} Moving on to the impact of the first stage, Table~\ref{tab:pvsg} reveals that the end-to-end VPS model appears to lag behind the IPS+T baselines. While VPS models have demonstrated their effectiveness on established datasets like Cityscape-VPS and KITTI-STEP, the PVSG dataset, characterized by its longer and more dynamic videos with frequent and significant shifts in camera view, presents novel obstacles for VPS research. This is evident in Figure~\ref{fig:exp_vis}, where the VPS models' subpar tracking ability significantly hampers their PVSG task performance. Table~\ref{tab:pvsg} also reflects this, particularly at a 0.1 threshold, where a minimal overlap in masks is sufficient for a recall. Here, the VPS results are nearly on par with IPS+T at R/mR@20, indicating that when the criteria for mask tube overlap are less strict, VPS can almost reach IPS+T levels, though not quite.

\paragraph{Understanding Numbers.} When examining Table~\ref{tab:pvsg}, it is crucial to prioritize the R/mR@20 as it represents our most significant indicator. The highest value for R@20 currently stands at 3.88, meaning that roughly for every 25 ground-truth triplets, one meets the criteria for a successful recall, indicating a relatively low efficiency. However, when setting the threshold to 0.1, the score improves to around 10, meaning the model can predict one in every 10 triplets with a looser requirement of recall. This suggests that while the model has some capability in recognizing key video content, there is substantial room for improvement in its accuracy and effectiveness.

%% file: section/7_discussion.tex
\section{Conclusion, Challenges, and Outlook}
\label{sec:discussion}
In this paper, we introduce a new PVSG task, a new PVSG dataset with several baselines to address the new task, in the hope of encouraging comprehensive video understanding and triggering more interesting downstream tasks such as visual reasoning. Here we discuss the challenges and future work.

\paragraph{Challenges}
Real-world data often exhibit long-tailed distributions across objects and relations, as shown in Figure~\ref{fig:pvsg_stat}. The PVSG models are expected to predict informative and diverse relations, rather than being obsessed with statistically common relations. Yet another challenge the PVSG models face is the uncertainty in relation descriptions. For example, ``playing with'' can be overlapping with ``chasing'' when it describes two kids chasing each other. Another important challenge is that the PVSG models seem to largely rely on video panoptic segmentation. With the video with a large view shift, the VPS models are expected to have a better performance on tracking and segmentation. Additionally, the time span prediction is a critical aspect of the PVSG model. A more sophisticated time-series technique could benefit the model development.

\paragraph{Outlook on Video Perception and Reasoning}
We foresee the potential of PVSG in bridging video scene perception and reasoning. While current video question-answering datasets lack pixel-level segmentation masks that refine (sometimes determine) the relations between object pairs, the inclusion of such dense annotations could be critical to video reasoning tasks. In fact, the PVSG dataset also provides dense captioning and question-answering annotation for each video, which could benefit the topic of reasoning and conversational chatbots.
PVSG is related to social intelligence, with rich event annotations in human behaviors and dynamics. In this spirit, the PVSG models might be critical to embodied agent tasks or virtual reality techniques, as the egocentric data is especially highlighted in the dataset.

\paragraph{Potential Negative Societal Impacts}
This work releases a dataset containing human behaviors, posing possible gender and social biases inherently from data. Potential users are encouraged to consider the risks of overlooking ethical issues in imbalanced data, especially in underrepresented minority classes.

\paragraph{Acknowledgement}
This study is supported by the Ministry of Education, Singapore, under its MOE AcRF Tier 2 (MOE-T2EP20221-0012), NTU NAP, and under the RIE2020 Industry Alignment Fund – Industry Collaboration Projects (IAF-ICP) Funding Initiative, as well as cash and in-kind contribution from the industry partner(s). We are also grateful to SuperAnnotate\footnote{\url{https://www.superannotate.com/}} for providing an outstanding annotation platform and excellent customer service. Special thanks to Binzhu Xie and Zitang Zhou from the Beijing University of Posts and Telecommunications for their dedicated leadership of the annotation team.

\paragraph{Author Contributions}
This paper represents a collaborative effort led by two principal contributors, \textbf{JY} and \textbf{WP}. \textbf{JY}, as the project leader, played a pivotal role in the initiative, monitoring and guiding each aspect, and managing both project and annotation teams, plus relation modeling development. \textbf{WP} significantly contributed by spearheading preliminary research, developing the PVSGAnnotation codebase, and crafting the video panoptic segmentation stage, which is fundamental to the OpenPVSG framework. The team was further supported by an experienced technical consultant, \textbf{XL}, on video segmentation details, and the academic consultants, \textbf{KZ} and \textbf{ZL} on project direction. Regular discussions and writing assistance were provided by \textbf{ZG}, \textbf{LC}, \textbf{BL}, and \textbf{MZ}. The project was further enriched by the invaluable resources and guidance from three mentors, \textbf{WZ}, \textbf{CCL}, and \textbf{ZL}, whose contributions were crucial to the project's success.

%% file: section/8_appendix.tex
\appendix
\setcounter{table}{0}
\renewcommand{\thetable}{A\arabic{table}}
\setcounter{figure}{0}
\renewcommand{\thefigure}{A\arabic{figure}}

\section{Implementation Details}
All experiments are performed in a single unified codebase called OpenPVSG, using the {\fontfamily{qcr}\selectfont MMDetection} framework to facilitate reproducibility.

\subsection{IPS+T}
\paragraph{Fine-tuning the Image Panoptic Segmentation Model} We first fine-tune the Mask2Former model (ResNet50 backbone) on the panoptic segmentation annotations from our PVSG dataset. Here we treat the PVSG dataset as an image dataset and process all video frames individually. The model is initialized from the best performing COCO-pretrained weights provided by {\fontfamily{qcr}\selectfont MMDetection} and then trained using a batch size of 32. The AdamW optimizer is used with a learning rate of 0.0001, weight decay of 0.05, and gradient clipping with a max L2 norm of 0.01. The learning rate is multiplied by 0.1 for the backbone, and the weight decay is set to 0.0 for embedding layers. Training runs for 8 epochs. 
\medbreak
\paragraph{Mask Association with Tracker} With both panoptic segmentation masks and the corresponding query features obtained from the fine-tuned Mask2Former model above, we then adopt the UniTrack model to associate masks in each frame to get the panoptic mask tubes and query feature tubes for each video clip. We configure the tracker using Unitrack's default config (config/imagenet\_resnet18\_s3\_womotion.yaml)\footnote{https://github.com/Zhongdao/UniTrack/} of Multi-Object Tracking and Segmentation (MOTS) setting and load pre-trained weights of their provided image-based SSL model {\fontfamily{qcr}\selectfont MoCoV1-ResNet50},  which has the best performance in MOTS task.

\subsection{VPS}
\paragraph{Fine-tuning the Video Panoptic Segmentation Model} We utilize Video K-Net implemented on a Mask2Former backbone as our VPS model, and train it using video panoptic segmentation annotations from our PVSG dataset. Optimal COCO-pretrained weights obtained from {\fontfamily{qcr}\selectfont MMDetection} are used to initialize the Mask2Former model.

\subsection{Relation Modeling}
\paragraph{Relation Dataset Formation} After completing the training of IPS+T and VPS models, we extract predicted feature tubes for each entity in the training videos. These tubes are segmented into individual frames for relation analysis. Specifically, within each frame, if both the subject and object predicted masks have a mask IOU greater than 0.5, we establish a relation between the pair. Subsequently, we map the relation annotations from ground truth pairs to these predicted pairs, forming the basis for the secondary stage of training.

\paragraph{Training} The training process begins with the use of two transformer models, one for the subject and one for an object, designed to encode each entity. This encoding ensures that each entity feature is enriched with contextual information from other entities in the frame. For the pair-selection model, we employ max pooling to distill the object feature tube into a singular object token. We then calculate cosine similarity to form a pairing matrix, indicative of potential relations. This matrix is contrasted with a ground truth matrix for supervision. In relation prediction, a multi-label loss computation is applied exclusively to pairs with established relations. The training is conducted with a batch size of 32, utilizing an Adam Optimizer with a learning rate of 0.001.

\section{Details of PVSG Formation}
In Section 4, we discuss several existing video datasets related to the PVSG dataset. In this section, we would like to elaborate on their characteristics and highlight our main considerations when building the PVSG dataset.

\medskip

\subsection{Video Selection and Focus}

We will first discuss some typical video datasets that are closely related to the PVSG task, and explain how we choose video sources to compose the PVSG dataset. 
We especially pay attention to the videos we think are better suited to explore contextual logic and reasoning. Here is a list of candidate datasets.

\noindent \textbf{Third-Person-View (TPV) video candidates} include VidSGG datasets (e.g., ImageNet-VidVRD~\cite{shang2017video}, VIDOR~\cite{shang2019vidor} and Action Genome~\cite{ji2020actiongenome}), video understanding dataset (e.g., TVQA~\cite{lei2018tvqa}, Howto100M~\cite{miech19howto100m}), and video panoptic segmentation dataset from VIP-Seg~\cite{miao2022large}.
     
\noindent \textbf{Egocentric video candidates} include Ego4D-STA~\cite{grauman2022ego4d}, VISOR~\cite{darkhalil2022epic}.

\medskip

\paragraph{ImageNet-VidVRD~\cite{shang2017video}: Short and Static Videos}
Although it is the first dataset to study video visual relations, most of VidVRD's video clips last only a few seconds and have almost static scenes. While it is a useful dataset for detecting relations between objects, additional research in logic and reasoning through temporal relation changes in videos may not be possible.

\begin{figure*}[!ht]
    \centering
    \includegraphics[width=\linewidth]{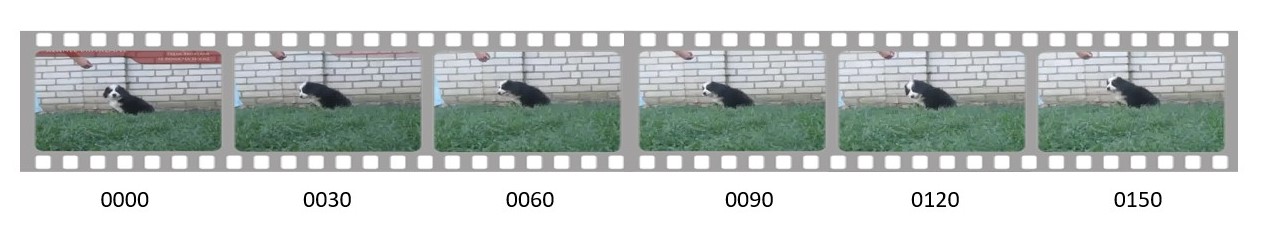}
    \caption{\textbf{An ImageNet-VidVRD example.} 
    The dog and the scene in this 7-second video barely change from start to finish.}
\end{figure*}

\paragraph{Action Genome~\cite{ji2020actiongenome}: Lack of Logic between Actions, Heavy Traces of Performance}
With dense relation annotations, Action Genome is commonly used in video scene graph generation task in recent years. Videos in Action Genome source from Charades dataset, which was made by 267 different users acting out certain sentences constructed by objects and actions from a fixed vocabulary. While some videos contain multiple actions and dynamic scenes, the transitions between these actions show heavy traces of performance. Due to the nature of the generated scripts, the whole video is more like a splicing of some instructed verbs. Consequently, there is no clear logic in the sequence of actions for observers to comprehend the video. In addition, simple videos with one or two actions made up a certain portion of this dataset. 
Therefore, we think Action Genome is not suitable for the PVSG task. 

\begin{figure*}[!h]
\label{fig:ag}
    \centering
    \includegraphics[scale=0.75]{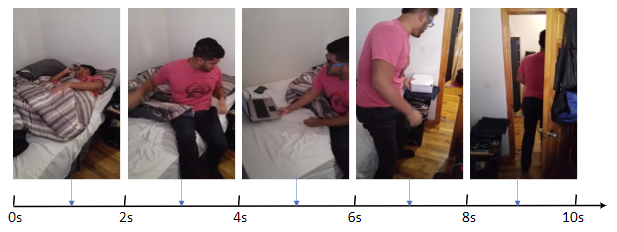}
    \caption{\textbf{An Action Genome example.} 
    In 10 seconds, the man quickly makes several actions: gets up from the bed, puts on his glasses, grabs and glances at the computer on the bed in 1 second, gets up and picks up the box, then walks out the door.}
\end{figure*}

\begin{figure*}[!ht]
    \centering
    \includegraphics[scale=0.6]{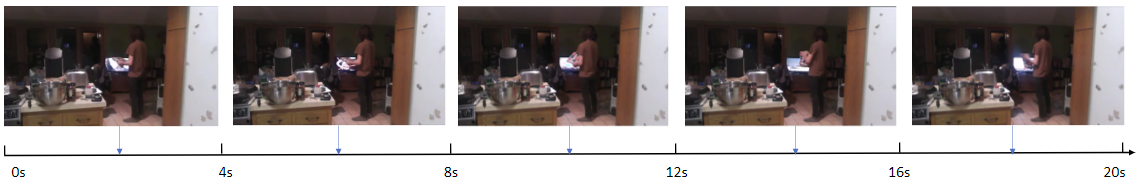}
    \caption{\textbf{Another Action Genome example.} 
    In 20 seconds, the man keeps standing in front of the TV and holding a book in his hand.}
\end{figure*}

\medskip

\begin{figure*}[!ht]
    \centering
    \includegraphics[scale=0.45]{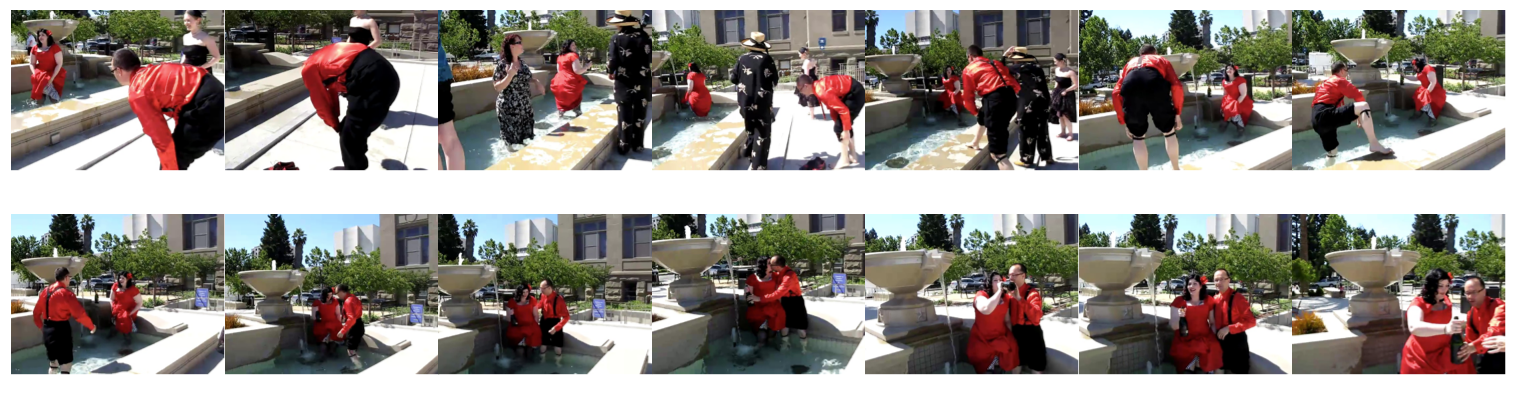}
    \caption{\textbf{A VIDOR good video example (demo video).} 
    At the beginning of the video, the man is pulling up his pants, the woman is holding up her skirt and standing in the water while the woman in black is taking pictures. We can guess from their outfits and the surroundings that this is a wedding photo shoot scene. Next, the man walks into the water and poses for photos with the woman. They hug each other, kiss each other, and drink liquor as the crowd cheers. We notice the sequence of actions in this one-minute demo video, with natural logical relationships between these actions. Both rich actions and coherent plots make this video more understandable and predictable. We think videos like this can make dynamic scene graphs better connected and attach more significance to the PVSG task.}
    \label{fig:vidor}
\end{figure*}

\paragraph{VIDOR~\cite{shang2019vidor}: A Good Candidate}
VIDOR is a large-scale dataset with all videos collected from user-uploaded videos on Flickr. Most of the videos are unedited records of daily life scenes, which ensures the coherence of the video plot and the natural connection of action changes. While there are useful videos in VIDOR, most of the videos contain too simple relations and some have ambiguous content. Therefore, we carefully select a subset of videos from VIDOR that fulfill our requirements to form the third-person view part of the PVSG dataset. The explanation of the good video in VIDOR is shown in Figure~\ref{fig:vidor}.
\subparagraph{Detailed Video Selecting Rules}
\begin{itemize}
 \item Selected videos should have main characters and contain a sequence of consistent actions (relations).
 \item Selected videos need to be comprehensible by observers.
 \item Discard videos with too many trivial and small objects for annotation purposes.
\end{itemize}

\medskip

\begin{figure*}[!h]
    \centering
    \includegraphics[width=\linewidth]{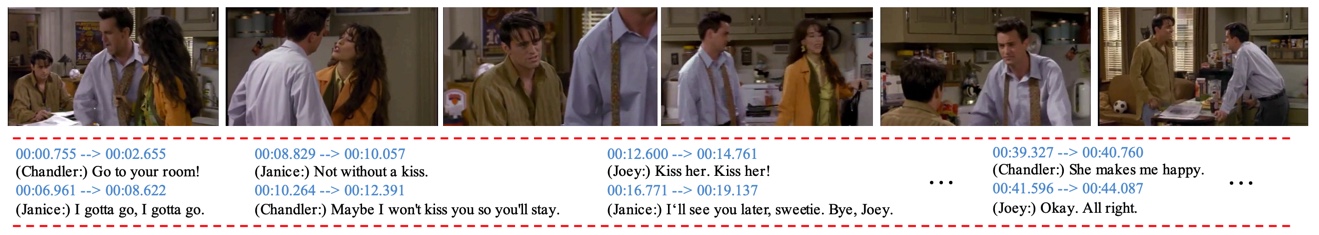}
    \caption{\textbf{TVQA example} 
    Cut shots appear every few seconds in TVQA videos.}
    \label{fig:tvqa}
\end{figure*}

\paragraph{TVQA~\cite{lei2018tvqa}: Too many cut shots}
TVQA is a frequently used dataset in Video Question Answering tasks. Since VQA also intends to study logical reasoning in videos, datasets in this domain lie in the relevant scope of the PVSG. However, videos from the TV show/movie datasets like TVQA contain lots of cut shots, which make it challenging to associate and relate objects across discontinuous scenes. Moreover, understanding such videos usually relies heavily on contextual information in the show. Hence, datasets like TVQA cannot apply to the PVSG task. Figure~\ref{fig:tvqa} shows an example for reference.

\medskip

\medskip

\begin{figure*}[!h]
    \centering
    \includegraphics[width=\linewidth]{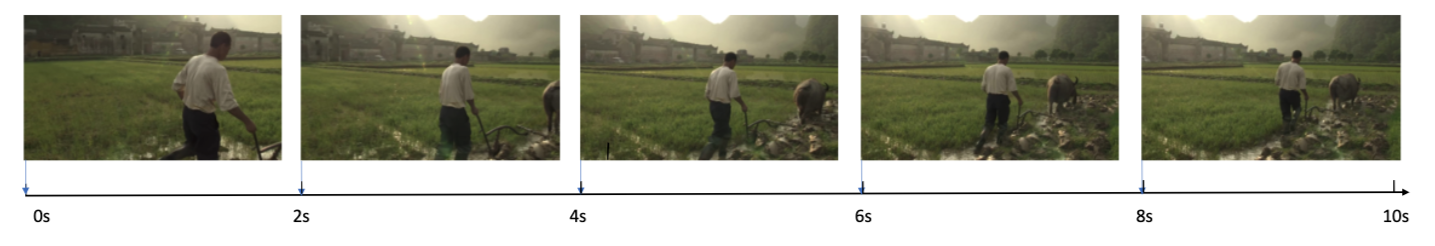}
    \caption{\textbf{A VIP-Seg example.} 
    The man keeps on plowing in this 10-second video clip.}
    \label{fig:vipseg}
\end{figure*}

\paragraph{VIP-Seg~\cite{miao2022large}: VPS but Static Videos}
VIP-Seg is a large-scale video panoptic segmentation dataset. Despite the fine annotations, the video scene is generally static and includes only one action due to the average video length of 5 seconds. Figure~\ref{fig:vipseg} shows an example for reference.

\begin{figure*}[!ht]
    \centering
    \includegraphics[width=\linewidth]{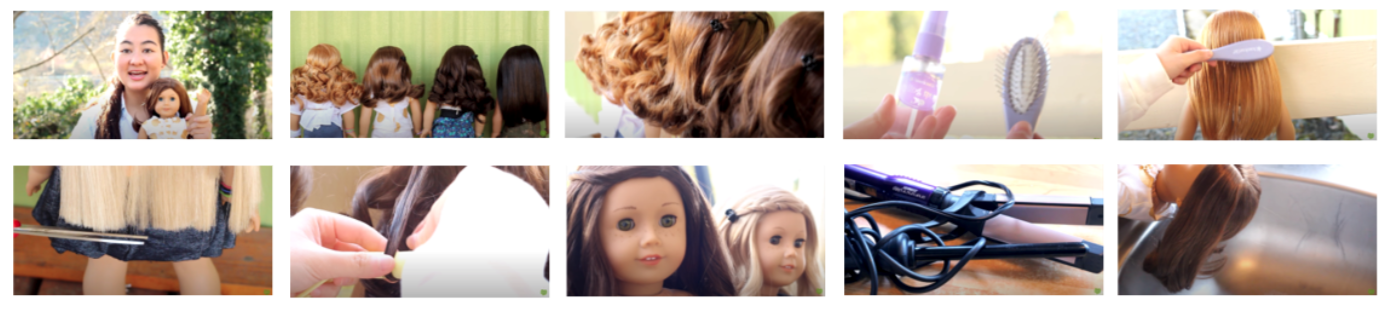}
    \caption{\textbf{An Howto100M example.} 
    A girl is teaching people how to care for American Girl Doll's hair.}
    \label{fig:howto}
\end{figure*}

\paragraph{Howto100M~\cite{miech19howto100m}: Curated videos with many cut shots}
HowTo100M is a large-scale dataset of instructional videos where complex tasks are broken into steps for the audience to learn. Clear logic can be found in such a dataset when a series of actions are made to address a specific task. However, one video contains many cut shots and is filmed from different angles (edited by the video creator). Hence, Howto100M does not fit for PVSG task. Figure~\ref{fig:howto} shows an example for reference.

\medskip

\begin{figure*}[!ht]
    \centering
    \includegraphics[width=\linewidth]{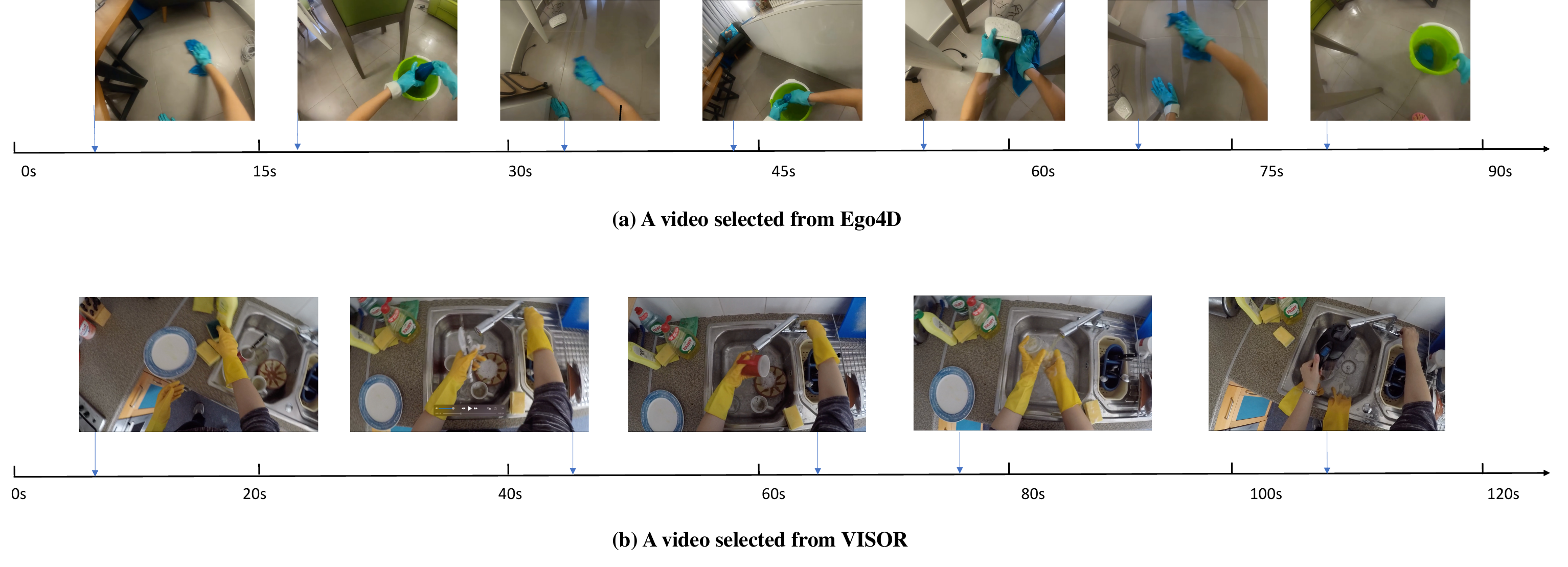}
    \caption{\textbf{Examples from Ego4D~\cite{grauman2022ego4d} and VISOR~\cite{damen2022epic}.} Both of these egocentric datasets are long videos (over 1 min), and the actors are doing specific tasks such as cooking and cleaning. Therefore, these videos inherently contain abundant contextual logic.}
    \label{fig:ego}
\end{figure*}

\paragraph{Egocentric Videos: Suitable for PVSG}
The PVSG dataset also contains egocentric videos, with the scope that the models should tackle the problems for both third-view videos and egocentric videos. Two high-quality egocentric datasets, Ego4D~\cite{grauman2022ego4d} and VISOR~\cite{darkhalil2022epic} are good candidates for selection. In fact, egocentric videos are usually taken when the actors are performing a specific task. The task usually contains a chain of actions with inherently clear logic. Therefore, egocentric videos are suitable for the reasoning task, and exploring scene graph generation based on these videos might invite a variety of techniques including perception and reasoning. Figure~\ref{fig:ego} shows an example for reference.

\begin{figure*}[!t]
    \centering
    \includegraphics[scale=0.3]{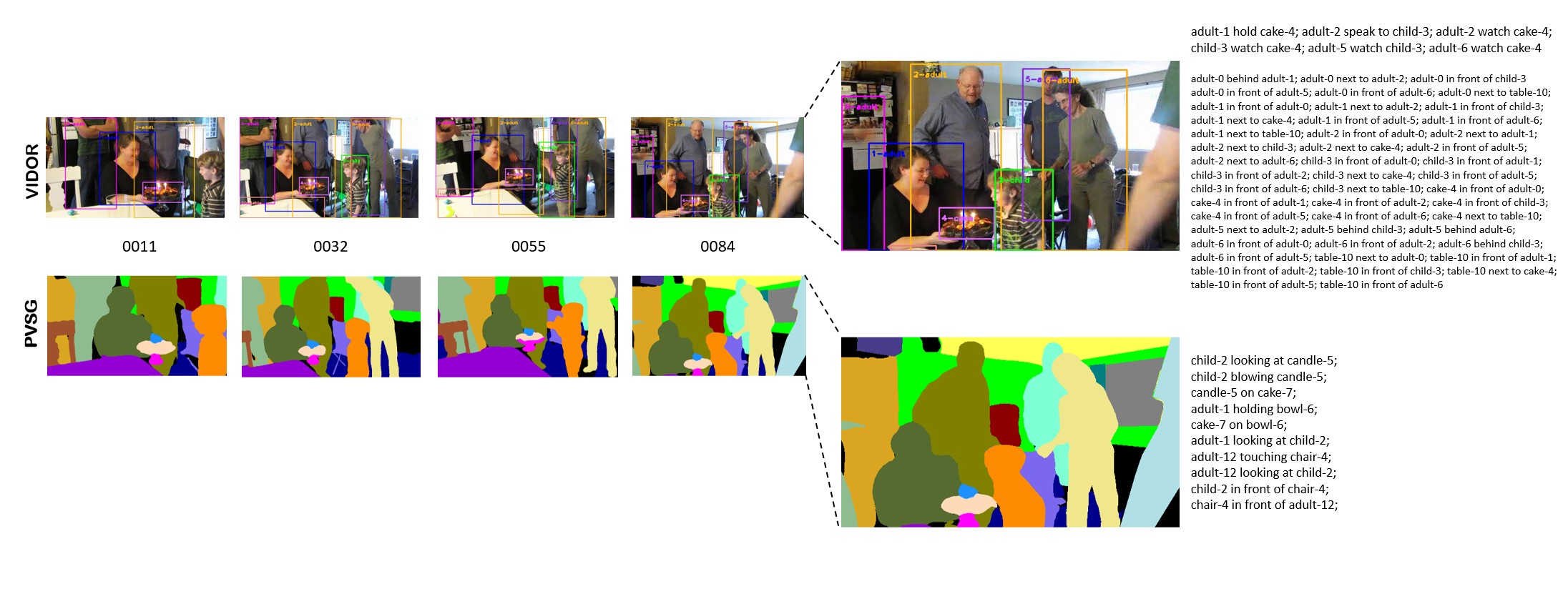}
    \caption{\textbf{Comparison between VIDOR and PVSG annotations.} 
    The VIDOR annotation uses bounding boxes to annotate objects in the videos. It is shown that some important objects are not annotated, such as candles. Without candles, important relations such as the kid blowing the candles can not be annotated too. In the PVSG dataset, the problem is solved by carefully defining the object classes. The high demand for our panoptic segmentation annotation also solves problems like the kid not being cropped out in Frame 0011. For relation annotation, the VIDOR contains many positional relations. However, most of the positional relations can be figured out in the static images, but the PVSG annotation highlights the dynamic relations in the video, 
    }
    \label{fig:vidor_annotation}
\end{figure*}

\subsection{Annotation Quality}
In this section, we will focus on our selected datasets discussed in Section B.1, compare their original annotations with the PVSG annotations, and analyze the necessity of panoptic segmentation for video scene graph generation and video reasoning.

\medskip

\paragraph{Comparison between VIDOR}
Figure~\ref{fig:vidor_annotation} shows the comparison between VIDOR and our PVSG dataset. We first observe the drawbacks of the VIDOR. 
1) It contains inconsistent bounding box annotations, e.g., the child is not cropped out at the first frame;
2) It misses important details, with no candle annotation, resulting in missing relations that are important to understand the scene. Actually, bbox is hard to annotate such small details;
3) There are many overlaps among different bounding boxes, one bounding box contains multiple object features;
4) Relation annotations only have simple predicates such as "watch", "hold" and many prepositions in the example frame, which cannot really describe what is happening in this frame;
5) Bounding box cannot annotate stuff such as water and ground. Actually, they also play an important role to understand the scene.

With all the considerations above, we annotate the PVSG dataset with great caution, such as 1) having a consistent annotation for all the objects. Once they are decided to be annotated, they will be annotated throughout the video; 2) we carefully design the object vocabulary beforehand after watching all the videos in the dataset to ensure all important objects are annotated; 3) panoptic segmentation avoids overlapping masks; 4) when annotating the relations, we focus on the dynamic relations rather than having many positional relations; 5) Panpotic segmentation is able to annotate the background that are critical to comprehensive scene understanding.